\documentclass[sigconf,natbib=false]{acmart}
\AtBeginDocument{%
  }

\setcopyright{acmlicensed}
\copyrightyear{2026}
\acmYear{2026}
\acmDOI{XXXXXXX.XXXXXXX}
\acmConference[SIGSPATIAL '26]{The 5th International Workshop on Spatial Big Data and AI for Industrial Applications}{Nov 03 2026}{Riverside, CA}

\acmISBN{978-1-4503-XXXX-X/2018/06}

\usepackage{listings}
\usepackage{xcolor}

\RequirePackage[
  datamodel=acmdatamodel,
  style=acmnumeric,
  ]{biblatex}
\definecolor{codegreen}{rgb}{0,0.6,0}
\definecolor{codegray}{rgb}{0.5,0.5,0.5}
\definecolor{codepurple}{rgb}{0.58,0,0.82}
\definecolor{backcolour}{rgb}{0.95,0.95,0.92}

\lstdefinestyle{codestyle}{
    backgroundcolor=\color{white},
    commentstyle=\color{codegreen},
    keywordstyle=\color{magenta},
    numberstyle=\tiny\color{codegray},
    stringstyle=\color{codepurple},
    basicstyle=\ttfamily\footnotesize,
    breakatwhitespace=false,
    breaklines=true,
    captionpos=b,
    keepspaces=true,
    numbers=left,
    numbersep=5pt,
    showspaces=false,
    showstringspaces=false,
    showtabs=false,
    tabsize=2
}
\copyrightyear{2026}
\acmYear{2026}
\setcopyright{cc}
\setcctype{by}
\acmConference[GeoIndustry '26]{The 5th ACM SIGSPATIAL International Workshop on Spatial Big Data and AI for Industrial Applications}{November 03--06, 2026}{Riverside, CA, USA}
\acmBooktitle{The 5th ACM SIGSPATIAL International Workshop on Spatial Big Data and AI for Industrial Applications (GeoIndustry '26), November 03--06, 2026, Riverside, CA, USA}
\acmDOI{10.1145/3849727.3856796}
\acmISBN{979-8-4007-3053-5/2026/11}

\begin{document}

\title{Trustworthy Geospatial AI Workflows: Reproducible Maritime Analytics and Forecasting on Open-Source AIS Data}


\author{Divya Acharya}
\affiliation{%
 \institution{Simula Research Laboratory}
 \city{Oslo}
 \country{Norway}}
\email{divya@simula.no}

\author{Thomas M. Roehr}
\affiliation{%
 \institution{Simula Research Laboratory}
 \city{Oslo}
 \country{Norway}}
 \email{roehr@simula.no}
 
\author{Arnaud Gotlieb}
\affiliation{%
 \institution{Simula Research Laboratory}
 \city{Oslo}
 \country{Norway}}
\email{arnaud@simula.no}



\renewcommand{\shortauthors}{Acharya et al.}

\begin{abstract}

Publicly available geospatial data provide increasing opportunities for data-driven maritime analytics and predictive modelling. However, their
availability does not necessarily imply that the underlying observations
are equally suitable for downstream AI applications. Properties of the
data source and observation process, including provenance, reporting
regularity, spatial coverage, observation density, and missingness, can
affect how geospatial data are interpreted, processed, and ultimately
used by predictive models. At the same time, trustworthy geospatial AI
requires analytical workflows in which these processing decisions can
be made explicit and reproduced.
This paper presents a reproducible maritime geospatial AI workflow built
around \emph{damast}, an open-source data-processing and machine-learning
library developed at Simula Research Laboratory. Using AIS observations from multiple public data providers as a case study, we demonstrate metadata-aware validation, reusable processing pipelines, maritime trajectory construction, and reproducible machine-learning experimentation. The predictive stage is model-agnostic and is instantiated in this study using a Transformer-based vessel-trajectory forecasting model.
Building on these implemented capabilities, we further outline an
envisioned extension toward trustworthiness-aware geospatial AI. The
extension introduces explicit characterization of source- and
data-level properties and their relationship with downstream model
performance. This distinguishes capabilities demonstrated in the
current workflow from trustworthiness-oriented components identified
for future development. Overall, the
study demonstrates how reproducible open-source infrastructure can
provide a practical foundation for developing more transparent and
trustworthy geospatial AI workflows.
\end{abstract}




\maketitle

\section{Introduction}
\label{sec:introduction}

The growing availability of publicly accessible geospatial data is creating new opportunities for data-driven applications across transportation, environmental monitoring, urban systems, and maritime operations \cite{2024:Wu:GeospatialBigData,2025:Dritsas:RemoteSensingGeospatial}. Such data enable researchers and practitioners to develop spatial analytics and artificial intelligence (AI) solutions without relying exclusively on proprietary data infrastructures. In the maritime domain, the Automatic Identification System (AIS) represents an important source of spatiotemporal information, providing observations of vessel positions and movements together with attributes such as speed, course, navigation status, and vessel type \cite{ITU-R-M1371-6}. The increasing accessibility of AIS data therefore provides a valuable foundation for maritime traffic analysis, vessel trajectory modelling, situational awareness, and predictive applications \cite{2018:Tu:ExploitingAISDatab}.

However, the public availability of geospatial observations does not necessarily imply that the resulting data are spatially and temporally homogeneous or equally suitable for downstream AI applications. AIS observations can exhibit irregular reporting intervals, temporal gaps, uneven spatial coverage, and differences in observation characteristics across vessels and geographical areas \cite{2023:Szarmach:ReconstructingDamagedData,2026:Jiang:ReviewVesselTrajectory}. These characteristics can influence derived trajectory features and consequently affect models trained on the resulting sequences. Nevertheless, geospatial AI pipelines often focus primarily on model performance after preprocessing, while the relationship between characteristics of the underlying public observations and the reliability of downstream predictions receives comparatively less attention.

This issue is particularly relevant for vessel trajectory forecasting. Recent advances in deep learning, including attention-based and Transformer architectures, provide powerful mechanisms for modelling complex spatiotemporal dependencies in vessel trajectories \cite{2023:Li:AISDatadrivenShip,2025:Yu:AISFormerLongtermVessel}. Forecasting performance, however, depends not only on model architecture but also on the observations from which trajectory sequences are constructed. Irregular sampling, reporting gaps, and geographically heterogeneous observation density may affect the information available to a forecasting model. Consequently, evaluating predictive performance independently of the characteristics of the underlying observation process can provide an incomplete view of model reliability.

At the same time, analysing large-scale public geospatial data requires processing workflows that are reproducible, extensible, and capable of preserving information about data semantics throughout the analytical lifecycle. To support these requirements, we develop and maintain \emph{damast}~\cite{Roehr:2026:damast}, an open-source framework for reusable data processing and machine-learning experimentation. \emph{damast} associates metadata with data frames, supports consistency checks over data properties such as units and value ranges, and enables the construction of reusable transformation pipelines. Its extensible architecture further allows domain-specific processing functionality to be incorporated while maintaining a common processing and experimentation infrastructure. In the maritime domain, these capabilities support trajectory-oriented feature engineering, spatial aggregation, reporting-gap analysis, and machine-learning forecasting within a reproducible workflow.

This paper investigates how reproducible geospatial AI workflows can
provide a foundation for more trustworthy use of public maritime data.
Using publicly available AIS observations from
BarentsWatch\footnote{\url{https://www.barentswatch.no/en/}} and Marine Cadastre\footnote{https://hub.marinecadastre.gov/pages/vesseltraffic}, we
demonstrate a \emph{damast}-based workflow covering metadata-aware
processing, harmonization of data from multiple sources, trajectory representation, sequence construction, and
predictive modelling, instantiated with a Transformer-based
vessel-trajectory temporal forecasting model. Building on this implemented
workflow, we outline an envisioned extension that incorporates explicit
source and data characterization and trustworthiness-aware evaluation.
We distinguish these envisioned components from those implemented in
the present study, providing a path toward geospatial AI workflows that
more explicitly account for the characteristics and limitations of
public data sources.

Taken together, the implemented and envisioned components connect
three aspects that are commonly considered separately: reproducible
geospatial data processing, characterization of public spatiotemporal
observations, and predictive AI. The current \emph{damast}-based
workflow provides the reproducible processing and predictive modelling
foundation, while the envisioned extensions introduce explicit
characterization of source and data properties and their connection
to downstream model evaluation. This broader perspective provides a
path toward geospatial AI workflows in which predictive performance
can ultimately be interpreted not only through aggregate accuracy
metrics, but also in the context of the public observations from which
the predictions are derived.
The contributions of this paper are as follows:
\begin{itemize}
    \item We present a reproducible maritime geospatial AI workflow and
outline its envisioned extension toward trustworthiness-aware AI,
distinguishing between currently implemented capabilities and
source- and data-characterization components identified for future
development.
    \item We demonstrate how \emph{damast}, our open-source framework,
supports metadata-aware validation, extensible and reusable
data-processing pipelines, and reproducible machine-learning
experimentation for public maritime AIS data.

\item We identify relevant characteristics of public AIS observations across spatial regions, temporal reporting patterns, and vessel categories, and describe how these characteristics can be represented within the proposed trustworthiness-aware extension.

    \item We apply the workflow to publicly available AIS
observations as a concrete maritime case study, demonstrating the
reproducible path from public observations through trajectory
representation and sequence construction to predictive modelling.
\end{itemize}


\section{Background and Related Work}
\label{sec:related-work}

Maritime geospatial AI spans several research directions, including
public vessel-tracking data, AIS trajectory analysis, trajectory
forecasting, and reproducible spatio-temporal data processing.
This section reviews these areas with particular attention to how
public AIS observations and their processing support downstream
predictive modelling.
\subsection{Public Geospatial Data for Maritime Analytics}
\label{subsec:public-geospatial-data}

The increasing availability of geospatial data enables large-scale
analysis across transportation, mobility, and environmental monitoring.
In the maritime domain, AIS provides time-stamped vessel positions and
navigational attributes that form spatio-temporal trajectory datasets
supporting traffic analysis, vessel behaviour modelling, and trajectory
prediction \cite{2018:Tu:ExploitingAISDatab}.

Public-sector infrastructures further increase access to maritime data.
In Norway, BarentsWatch provides AIS vessel positions collected by the
Norwegian Coastal Administration through APIs and downloadable datasets
\cite{barentswatch_open_data}, supporting reproducible research without
exclusive reliance on proprietary sources. Denmark likewise publishes datasets~\footnote{http://aisdata.ais.dk} for its Exclusive Economic Zone (EEZ), and Marine Cadastre regularly publishes coastal AIS data from the US.  Together, these public infrastructures illustrate that AIS datasets are
available across different geographical and institutional settings, while their formats, coverage, and observation characteristics may differ, motivating reproducible workflows that can harmonize heterogeneous public AIS sources. However, accessibility does not imply spatially or temporally uniform observations; AIS trajectory characteristics may vary across space, time, and vessel operating conditions \cite{2026:Jiang:ReviewVesselTrajectory}. Such variation is relevant when observations are transformed into inputs for downstream geospatial AI models.

\subsection{AIS Data Quality and Maritime Trajectory Analysis}
\label{subsec:ais-quality}

Raw AIS observations do not constitute uniformly sampled vessel trajectories.
Transmission intervals vary with vessel movement and operating conditions,
while signal loss, receiver coverage, and communication constraints may
introduce temporal gaps. AIS datasets may additionally contain missing or
erroneous attributes, duplicated observations, positional outliers, and
kinematically inconsistent records
\cite{2023:Szarmach:ReconstructingDamagedData,2026:Jiang:ReviewVesselTrajectory}. Consequently, AIS
processing commonly involves filtering, trajectory segmentation,
interpolation or resampling, and the derivation of movement-related features
before subsequent analysis.

These processing decisions influence the trajectory representation supplied
to downstream models. Recent work has shown that reconstruction and
preprocessing strategies can affect vessel trajectory prediction,
particularly for sparse or irregular observations
\cite{2026:Goh:DatadrivenAISTrajectory}. AIS characteristics can also vary
geographically: differences in traffic activity and observation conditions
produce regions with different message densities, numbers of observed
vessels, and reporting patterns.

These characteristics are relevant not only as preprocessing concerns,
but also as contextual information about the observations from which
trajectory representations are constructed. Reporting intervals,
trajectory gaps, spatial observation patterns, and vessel-related
characteristics therefore motivate a broader view of AIS data quality
in which properties of the observation process remain visible when
designing and evaluating downstream geospatial AI workflows. Explicitly
incorporating such characteristics into trustworthiness-aware model
evaluation forms part of the envisioned extension discussed later in
this paper.

\subsection{Vessel Trajectory Forecasting}
\label{subsec:trajectory-forecasting}

Vessel trajectory forecasting estimates future positions from
historical movement observations, supporting traffic management,
navigational safety, and decision support. While earlier approaches
relied largely on kinematic and state-estimation methods, the
availability of AIS data has encouraged data-driven models capable
of learning nonlinear movement patterns
\cite{2023:Li:ShipTrajectoryPrediction}.
Recurrent architectures such as LSTM and GRU have been widely applied
to sequential AIS observations
\cite{2025:Slaughter:VesselTrajectoryPrediction,2024:Zaman:EfficientLongShortTerm}.
More recently, Transformer-based approaches use self-attention to
model relationships across trajectory observations and have been
extended through alternative attention mechanisms, feature
representations, and hybrid components
\cite{2024:Wang:TPTransVesselTrajectory,2026:Lee:AISDatabasedShip}.
Forecasting performance nevertheless depends on trajectory
construction, input representation, prediction horizon, geographical
context, and properties of the underlying AIS observations
\cite{2026:Lee:AISDatabasedShip}. Accordingly, the Transformer in
this work is not proposed as a new architecture, but serves as the
predictive component of the reproducible \emph{damast}-based
workflow. Its model-agnostic design allows alternative predictive
architectures while retaining the same data-processing and
trajectory-construction stages.
\subsection{Reproducible Geospatial AI}
\label{subsec:reproducible-geospatial-ai}

Reproducibility in geospatial AI extends beyond preserving a trained
model to include data semantics, processing and feature construction,
model configuration, and evaluation procedures. FAIR-oriented and
scientific-workflow research therefore emphasizes explicit
representation of data, processing steps, provenance, and
computational artifacts to support transparency and reuse
\cite{2020:Goble:FAIRComputationalWorkflows,2022:Barker:IntroducingFAIRPrinciples}.

This is particularly relevant to maritime trajectory analysis, where
temporal segmentation, movement features, spatial aggregation, and
sequence construction determine the representations supplied to
forecasting models
\cite{2026:Jiang:ReviewVesselTrajectory,2020:Chen:ShipTrajectoryReconstruction}.
Reproducible maritime geospatial AI should therefore preserve both
processing steps and the semantics of the variables and
transformations involved.

Existing research has advanced AIS data-quality analysis, vessel trajectory forecasting, and reproducible geospatial workflows, but these concerns are commonly addressed separately.  The contribution of this work is not a new AIS preprocessing operation or forecasting architecture. Rather, it integrates metadata-aware validation, serializable geospatial processing, trajectory and sequence construction, and model experimentation within a common reproducible workflow, while providing a foundation for explicit consideration of source and data characteristics. This motivates the \emph{damast}-based workflow
demonstrated in this study and the trustworthiness-oriented
extensions envisioned in the following sections.

\section{\emph{damast}: An Open-Source Framework for Reproducible Geospatial AI}
\label{sec:damast}

The open-source Python library \emph{damast} \cite{Roehr:2026:damast},
developed at Simula Research Laboratory since 2023, supports reproducible data
processing and machine-learning experimentation, with maritime AIS
time-series analysis and anomaly detection as initial motivating
applications. Its core objective is to make domain knowledge about
data and its semantics explicit, including units, specifications, and representation
types. By embedding shareable metadata in its preferred Parquet
storage format, \emph{damast} facilitates the understanding,
reproduction, and reuse of data-processing and modelling workflows,
while enabling subsequent processing stages to establish explicit
interface contracts.

The design of \emph{damast} is motivated by three recurring reproducibility challenges in data-driven research:
(1) understanding the properties and semantics of the datasets,
(2) reconstructing how processing pipelines transform raw observations into analysis-ready or model-ready representations, and
(3) reproducing how machine-learning models are configured, trained, and evaluated. 
These challenges become especially relevant for public geospatial data, where heterogeneous formats with inconsistent naming schemas, undocumented assumptions, derived attributes, and large data volumes can make apparently simple analytical workflows difficult to reproduce.
The library \emph{damast} addresses these challenges through three closely connected layers: \emph{data understanding}, \emph{reproducible processing}, and a \emph{machine-learning experiment framework} to focus the machine-learning model design. Together, these layers provide a traceable path from the original observations to derived geospatial features and predictive models.

\subsection{Design and Architecture}
\label{subsec:damast-architecture}

At the data level, \emph{damast} associates tabular observations with explicit metadata describing the structure and semantics of the dataset. Metadata may include column specifications, admissible value ranges, units, descriptive statistics, provenance, licensing information, and additional annotations. The goal is to ensure that the meaning of individual variables does not remain implicit in application code.

This is particularly important for geospatial and maritime data. An AIS dataset may contain latitude, longitude, timestamps, vessel identifiers, speed, course, and other attributes, but subsequent analysis frequently introduces additional variables whose interpretation depends on the transformations used to create them. For example, a vessel speed estimate may be derived from spatial displacement and elapsed time rather than being directly observed. Reproducing such a quantity requires knowledge of the distance definition, temporal reference, units, filtering assumptions, and ordering of observations.

\emph{damast} therefore combines data values and metadata through its annotated data abstractions. Processing operations can explicitly describe the attributes they consume and produce, including semantic requirements such as expected units and ranges. These specifications provide consistency checks between processing stages and make assumptions about the input data visible within the pipeline rather than leaving them embedded in implementation details.

The library also provides a command-line interface with sub-commands addressing common stages of the data lifecycle: (\texttt{inspect}) enables the examination of a dataset and metadata properties, including the possibility to inspect compressed files such as zip-files of zip-files, filtering on values and ranges, (\texttt{annotate}) supports creation and modification of dataset metadata, e.g., to set units, description and correcting the representation type, (\texttt{convert}) provides conversion between commonly used storage formats, e.g., from CSV to Parquet, and (\texttt{process}) permits to run previously defined processing pipelines, stored as yaml artifacts, to compatible datasets.
This makes \emph{damast} usable both programmatically and through  command-line based workflows.

\subsection{Reproducible and Extensible Data Processing}
\label{subsec:damast-processing}

A central concept in \emph{damast} is that reproducibility requires preserving the complete sequence of transformations applied to a dataset. Rather than representing preprocessing as an informal collection of scripts or notebook cells, \emph{damast} models transformations as explicitly defined pipeline elements that can be connected to form a \texttt{DataProcessingPipeline}.
This design builds upon the pipeline and transformer concepts used in scikit-learn~\footnote{\url{https://scikit-learn.org/stable/}}, while extending them with metadata-aware annotation, input validation, and an explicit processing graph. 
The focus of \emph{damast} is on maintaining semantic consistency across processing pipelines and bridging the gap between research and production pipelines. Therefore, unlike Apache Spark~\footnote{https://spark.apache.org}, \emph{damast} does not target large scale distributed workflows for production.

Designers of individual pipeline transformations can define their required input columns and generated output columns. Where applicable, these specifications also include units, ranges, or other metadata constraints. This makes it possible to identify inconsistencies before they propagate into downstream analysis or machine-learning stages.
For AIS processing, this distinction is particularly important because a substantial proportion of the variables used in trajectory analysis are derived rather than directly observed. Starting from vessel identifiers, timestamps, and geographical coordinates, a processing pipeline may derive inter-message time differences, travelled distance, estimated speed, heading, angular velocity, cyclic temporal variables, and spatial cell identifiers. The interpretation of each derived quantity depends on the preceding processing operations and their associated assumptions.
\begin{figure*}[]
    \centering
    \includegraphics[width=0.85\textwidth]{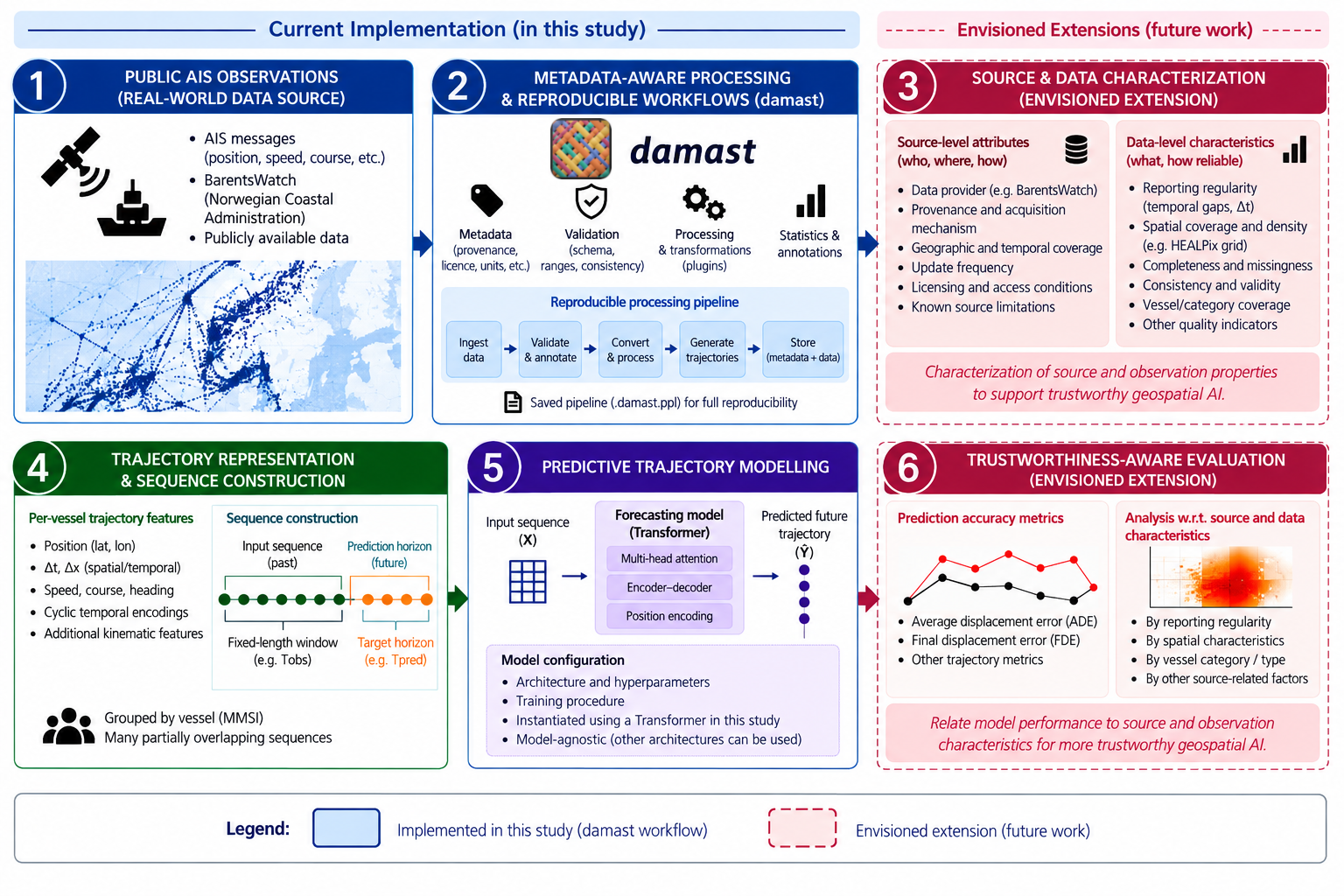}
    \caption{Implemented and envisioned stages toward trustworthy maritime
    geospatial AI. Stages 1, 2, 4, and 5 are implemented using reproducible
    \emph{damast} pipelines and Transformer-based forecasting; Stages 3 and 6
    represent envisioned source/data characterization and
    trustworthiness-aware evaluation.}
    \label{fig:source-aware-framework}
\end{figure*}
\subsection{Reproducibility Across the Geospatial AI Lifecycle}
\label{subsec:damast-reproducibility}
In our implementation, \emph{damast} is extended with maritime-specific transformations while retaining the generic processing infrastructure. Custom transformations are discovered through the \emph{damast} plugin mechanism, and can therefore be incorporated into saved processing pipelines without modifying the core framework.
This enables domain-specific functionality to evolve independently while maintaining a common mechanism for validation, execution, and pipeline serialization. 

For the AIS analysis used in this work, the preprocessing workflow includes timestamp parsing, filtering, temporal-difference calculation, distance and kinematic feature generation, cyclic transformations, and HEALPix-based~\cite{2005:Gorski:HEALPixFrameworkHighResolution} spatial binning. The resulting pipeline can be serialized as a \texttt{.damast.ppl} file and subsequently replayed on compatible datasets using the generic \emph{damast} processing interface. This allows the sequence of transformations used to produce an analytical dataset to be retained as an explicit, reusable research artifact.
Steps that might have side-effects can also be marked, so that the existence of this artifacts after the step execution can be checked for.

The maritime extension further supports aggregation of vessel-level and fleet-level reporting statistics. In particular, reporting intervals between consecutive AIS observations can be summarized using statistics such as their mean, median, standard deviation, minimum and maximum values, together with the proportions of observations exceeding predefined temporal-gap thresholds. These statistics provide descriptive information about the temporal reporting characteristics of the public AIS observations.
Spatial characteristics can likewise be represented within the
processing workflow. AIS observations are mapped into hierarchical
HEALPix cells, after which reporting behaviour, vessel activity, and
vessel-category composition can be summarized geographically. This
allows spatial areas with comparatively sparse or irregular reporting
to be identified without treating the entire AIS dataset as a
homogeneous observation source. These descriptive capabilities provide
a technical basis for the more explicit source- and data-characterization
stage envisioned in Section\ref{sec:framework}.

\subsection{Machine Learning and Forecasting}
\label{subsec:damast-ml}

Reproducibility becomes more challenging when processed data are subsequently used for machine learning. In trajectory forecasting, for example, the final result depends not only on the selected model architecture, but also on the construction of vessel-specific sequences, the historical sequence length, forecasting horizon, feature representation, preprocessing configuration, training parameters, and model implementation.

\emph{damast} addresses this problem by extending the processing abstraction to machine-learning experimentation. Its experiment layer is designed to associate the preprocessing workflow with model configurations, learning-task definitions, training parameters, and experiment artifacts. This provides a common structure in which alternative predictive models can be evaluated over identically processed observations.

For sequential prediction, \emph{damast} predefines an abstraction named \texttt{TemporalForecastTask}. Observations can be grouped according to an entity identifier, such as the Maritime Mobile Service Identity (MMSI), to construct temporally ordered vessel sequences. Let

\begin{equation}
    \mathbf{X}^{(v)}_{H^{-}}
    = \sigma
    \left(\left[
    \mathbf{x}^{(v)}_{t-H^{-}+1},
    \ldots,
    \mathbf{x}^{(v)}_{t}\right], L
    \right)
\end{equation}

denote the historical trajectory of vessel $v$ over a temporal input horizon $H^{-}$ with samples interpolated onto equidistant time points, $\sigma$ as interpolation function, $L = |X^{(v)}_{H^{-}}|$ as sequence length and $\mathbf{x}^{(v)}_t$ contains the trajectory features at time $t$. A forecasting model $f_{\theta}$ estimates the subsequent trajectory over a temporal horizon $H^{+}$ as

\begin{equation}
    \hat{\mathbf{Y}}^{(v)}_{t+1:t+H^{+}}
    =
    f_{\theta}
    \left(
    \mathbf{X}^{(v)}_{H{-}}
    \right).
\end{equation}

The model architecture is separated from the data-processing and task specification. Consequently, multiple forecast model architectures can be evaluated using the same sequence construction, preprocessing, and evaluation configuration. In this study, we exploit this model-extensible interface to integrate a Transformer-based vessel trajectory forecasting model. 
In addition, \emph{damast} uses Keras \cite{chollet2015keras} as an abstraction layer for
machine-learning implementations, allowing different computational
backends, including PyTorch, JAX, and TensorFlow, to be used without
changing the higher-level experiment specification.

This separation also makes the forecasting stage model-agnostic, allowing different predictive architectures to be instantiated without modifying the preceding data-processing and trajectory-construction stages. 
The same data-processing pipeline can also be applied during inference,
while omitting transformations associated with target construction.
This ensures that prediction inputs undergo the same preprocessing
steps as the data used during model training, supporting reproducible
deployment and evaluation. In this study, we instantiate this capability using a Transformer-based trajectory forecasting
model.

Reproducibility and FAIR-oriented data processing are primary
motivations of \emph{damast}, which provides explicit representations
of data-processing pipelines and machine-learning experiments across
three connected levels: \emph{Data Understanding $\rightarrow$ Data
Processing $\rightarrow$ Model Experimentation}. Metadata describe
what observations represent, processing pipelines record how they
are transformed into derived features, and experiment configurations
capture how these features are used for predictive modelling.
Maintaining traceability across these levels is particularly relevant
for public geospatial AI, where upstream observation characteristics
and processing decisions may affect downstream model inputs. For
example, irregular AIS reporting intervals can influence derived
distance or speed estimates and consequently the trajectory sequences
supplied to forecasting models. The next section builds upon this reproducible infrastructure to
outline a broader vision for trustworthy maritime geospatial AI,
distinguishing between capabilities implemented in the current
\emph{damast} based workflow and source and data-aware components
envisioned for its future extension.

\section{Towards Trustworthy Maritime Geospatial AI Workflows}
\label{sec:framework}

The development of trustworthy geospatial AI requires consideration
of both the predictive model and the data-processing chain through
which model inputs are produced. In public maritime data, the prediction pipeline begins with an observation process that may be spatially and temporally heterogeneous, partial, and influenced by the
reliability of message publishers, reception infrastructure, and subsequent processing entities. Reporting frequency, spatial coverage,
observation density, and vessel-specific reporting patterns can
therefore influence the trajectory representations supplied to
predictive models.


The \emph{damast} framework developed in this work provides the
reproducible infrastructure for representing and executing this
analytical chain. As described in Section~\ref{sec:damast}, it combines
metadata-aware data handling, explicit and reusable processing
pipelines, extensible maritime transformations, trajectory and
sequence construction, and reproducible machine-learning
experimentation. These capabilities establish traceability between
the original observations, the representations derived from them,
and the predictive models operating on those representations. This
relationship can be expressed conceptually as
\begin{equation}
\small
 \text{Observed Data}
    \rightarrow
    \text{Derived Representation}
    \rightarrow
    \text{Predictive Output}.
\end{equation}

For AIS, these levels correspond respectively to the original vessel
observations, the trajectory and movement features derived through
processing, and the future vessel trajectory estimated by the
forecasting model.
Building on this implemented foundation, we propose an extension
toward trustworthiness-aware maritime geospatial AI in which
properties of both the data source and the observed data are made
explicit and considered in relation to downstream model behaviour.
At the source level, relevant properties may include provenance,
acquisition and reporting mechanisms, geographical and temporal
coverage, access conditions, and known limitations. At the data
level, relevant properties include reporting regularity, observation
density, missingness, spatial coverage, consistency, and variation
across vessels or vessel categories. Rather than treating such
properties as detached data-quality descriptors, the envisioned
extension places them within the same analytical life-cycle as
trajectory construction and predictive modelling.

The resulting perspective combines an implemented, reproducible
\emph{damast} based workflow with an envisioned trustworthiness-oriented
extension that connects source and data characteristics to downstream
model evaluation. Figure~\ref{fig:source-aware-framework} summarizes
this distinction between the components demonstrated in the present
study and those identified for further development.

\subsection{Framework Overview}
\label{subsec:framework-overview}

Figure~\ref{fig:source-aware-framework} summarizes the current
\emph{damast} based workflow together with the envisioned extensions
toward trustworthiness-aware maritime geospatial AI. The overall
workflow is organized into six stages:

\begin{enumerate}
    \item acquisition of publicly available AIS observations;
    \item metadata-aware validation and reproducible processing using \emph{damast};
    \item source and data characterization;
    \item trajectory feature representation and sequence construction;
    \item predictive trajectory modelling; and
    \item trustworthiness-aware evaluation of predictive outputs.
\end{enumerate}

Stages~1, 2, 4, and 5 constitute the workflow currently implemented and demonstrated in this study: public AIS observations are processed through metadata-aware, reproducible \emph{damast} pipelines, transformed into model-ready vessel trajectories, and supplied to a model-agnostic forecasting stage, here instantiated using a Transformer-based trajectory forecasting model.
Stages~3 and 6 represent envisioned extensions. Stage~3 introduces explicit characterization of source- and data-level properties; Stage~6 connects these characteristics to downstream model performance together supporting evaluation of predictive behaviour in light of the conditions under which the underlying observations were produced.



\subsection{Envisioned Source and Data Characterization}
\label{subsec:reliability-characterization}

The third stage represents an envisioned extension of the current
workflow in which characteristics of both the data source and the
resulting observations are represented explicitly. We distinguish
between \emph{source-level} and \emph{data-level} characteristics.
Source-level characteristics describe the context in which the data
are produced and made available, including provenance, acquisition
and reporting mechanisms, geographical and temporal coverage,
access conditions, and known limitations. Data-level characteristics
describe properties observable in the resulting dataset, including
reporting regularity, temporal gaps, observation density, missingness,
spatial coverage, consistency, and variation across vessels or vessel
categories.

The current \emph{damast} implementation already provides a
technical basis for deriving several data-level descriptors. For
example, reporting intervals and temporal-gap statistics can be
computed at vessel and fleet level, while HEALPix-based spatial
aggregation supports the analysis of geographical observation
patterns. In the envisioned extension, these descriptive capabilities
would be combined with explicit source-level metadata to provide a
more systematic characterization of the conditions under which
public AIS observations are produced and subsequently used for
predictive modelling.

To operationalize the data-level component of this envisioned stage,
we consider measurable descriptors of temporal reporting behaviour
and spatial observation patterns. These descriptors build directly
on processing capabilities already available in \emph{damast}, while
their systematic integration with source-level characteristics forms
part of the envisioned extension.

\paragraph{Temporal reporting characteristics.}
AIS observations are generated through an asynchronous reporting
process. Consequently, consecutive observations belonging to the
same vessel do not necessarily occur at uniform temporal intervals.
Let the ordered timestamps associated with vessel $v$ be
$
    t^{(v)}_1, t^{(v)}_2, \ldots, t^{(v)}_{N_v}.
$. 
The reporting interval between consecutive observations is defined as
\begin{equation}
    \Delta t^{(v)}_i =
    t^{(v)}_i - t^{(v)}_{i-1}.
\end{equation}
For each vessel, the distribution of $\Delta t$ can be summarized
using descriptive statistics including the mean, median, standard
deviation, minimum, and maximum reporting intervals. The proportion
of reporting gaps exceeding selected temporal thresholds provides an
additional descriptor of reporting continuity. Together, these
quantities characterize the temporal reporting behaviour of individual
vessel trajectories without interpreting reporting irregularity itself
as a direct measure of trustworthiness.

\paragraph{Spatial observation characteristics.}
To obtain a geographical view of the observation process, the study
area can be partitioned using the Hierarchical Equal Area isoLatitude
Pixelization (HEALPix) representation. Each AIS observation is assigned
to a spatial cell $h$. For each populated cell, descriptive quantities
such as observation density, number of unique vessels, average reporting
interval, proportion of long reporting gaps, and vessel-category
composition can be derived.

Let $\mathcal{O}_h$ denote the collection of AIS observations associated
with cell $h$. A cell-level reporting-gap descriptor can be expressed as
\begin{equation}
    G_h =
    \frac{1}{|\mathcal{O}_h|}
    \sum_{i \in \mathcal{O}_h}
    \Delta t_i.
\end{equation}
Larger values of $G_h$ indicate spatial cells in which consecutive
vessel observations are, on average, separated by longer temporal
intervals. Combined with observation density and vessel activity,
such descriptors provide a quantitative representation of spatial
variation in the observed AIS data.

These temporal and spatial descriptors are not treated as prediction
labels or direct measures of trustworthiness. Instead, they provide
measurable data-level characteristics that can be retained alongside
the processing and modelling workflow. In the envisioned extension,
these descriptors would be combined with source-level characteristics
and subsequently examined in relation to downstream model behaviour.
This provides a concrete path from the descriptive capabilities already
supported by \emph{damast} toward systematic trustworthiness-aware
evaluation.
\subsection{Maritime Trajectory Representation}
\label{subsec:trajectory-representation}
The fourth stage corresponds to the trajectory-representation and
sequence-construction component implemented within the current
\emph{damast}-based workflow. Raw AIS observations are converted into
a representation suitable for sequence-based forecasting. For each
vessel, observations are ordered temporally and grouped according to
MMSI. In addition to the original spatial coordinates, the processing
pipeline derives temporal and kinematic variables describing vessel
motion.

Let the geographical position of vessel $v$ at time $t$ be
\begin{equation}
    \mathbf{p}^{(v)}_t =
    \left(
    \phi^{(v)}_t,
    \lambda^{(v)}_t
    \right),
\end{equation}
where $\phi$ and $\lambda$ denote latitude and longitude, respectively.
The spatial displacement between consecutive observations is represented as
\begin{equation}
    \Delta d^{(v)}_t =
    d\left(
    \mathbf{p}^{(v)}_{t-1},
    \mathbf{p}^{(v)}_t
    \right),
\end{equation}

where $d(\cdot,\cdot)$ denotes the geographical distance function used in the processing pipeline.

Together with the corresponding temporal difference $\Delta t^{(v)}_t$, this quantity can be used to derive movement-related features such as estimated speed. Additional trajectory attributes include heading, angular velocity, and cyclic representations of temporal and spatial variables where appropriate.

Each vessel trajectory is then divided into fixed-length historical sequences, by interpolating existing samples onto equidistant time points. Let

\begin{equation}
    \mathbf{x}^{(v)}_t =
    \left[
    x^{(v)}_{t,1},
    x^{(v)}_{t,2},
    \ldots,
    x^{(v)}_{t,F}
    \right]
\end{equation}

denote the $F$-dimensional feature vector representing vessel $v$ at time $t$. A historical (interpolated) sequence over the temporal horizon $H^{-}$ of length $L$ is represented as

\begin{equation}
\mathbf{X}^{(v)}_{H^{-}}
    \in
    \mathbb{R}^{L \times F}.
\end{equation}

The corresponding forecasting target consists of the vessel positions over a temporal horizon $H^{+}$:

\begin{equation}
    \mathbf{Y}^{(v)}_{t+1:t+H^{+}}
    =
    \left[
    \mathbf{p}^{(v)}_{t+1},
    \ldots,
    \mathbf{p}^{(v)}_{t+H^{+}}
    \right].
\end{equation}

This formulation separates the density of the observed trajectory history from the future forecasting horizon $H^{+}$, allowing different experimental configurations to be evaluated within the same processing framework.

\subsection{Predictive Trajectory Modelling}
\label{subsec:predictive-modelling}

The fifth stage corresponds to predictive trajectory modelling within
the implemented \emph{damast}-based workflow. This stage is designed
to be model-agnostic, allowing alternative predictive architectures
to operate on the same reproducibly constructed trajectory
representations. The objective is not to introduce a new forecasting
architecture, but to demonstrate how predictive models can be
integrated into a reproducible geospatial AI workflow. In this study, we instantiate this component using a Transformer-based sequence model.


Given an input trajectory sequence
$\mathbf{X}=[\mathbf{x}_1,\mathbf{x}_2,\ldots,\mathbf{x}_L]$,
the feature vectors are first projected into a latent representation of dimension $d_{\mathrm{model}}$. Positional information is incorporated so that temporal ordering remains available to the attention layers.

For a sequence representation $\mathbf{Z}$, scaled dot-product attention \cite{2017:Vaswani:AttentionAllYou} is defined as \begin{equation}
    \mathrm{Attention}(\mathbf{Q},\mathbf{K},\mathbf{V})
    =
    \mathrm{softmax}
    \left(
    \frac{
        \mathbf{Q}\mathbf{K}^{T}
    }{
        \sqrt{d_k}
    }
    \right)
    \mathbf{V},
\end{equation}

where $\mathbf{Q}$, $\mathbf{K}$, and $\mathbf{V}$ denote query, key, and value representations, respectively.

Multiple attention heads allow the model to capture different relationships between trajectory observations. The resulting representation is passed through feed-forward layers and subsequently mapped to future geographical coordinates.
The forecasting model can therefore be written as
\begin{equation}
    \hat{\mathbf{Y}}
    =
    f_{\theta}(\mathbf{X}),
\end{equation}
where $\theta$ denotes the learned Transformer parameters.
The Transformer is implemented within the \emph{damast} forecasting experiment abstraction described in Section~\ref{subsec:damast-ml}. This enables sequence construction, feature selection, training configuration, and model evaluation to remain consistent across experiments and across alternative forecasting baselines. For the case study, the Transformer uses $d_{\mathrm{model}}=32$, four attention heads, two Transformer layers, a feed-forward dimension of 64, and a dropout of 0.1, with \texttt{latitude\_x}, \texttt{latitude\_y}, \texttt{longitude\_x}, and \texttt{longitude\_y} as input features. The Transformer therefore represents the predictive model instantiated in this study rather than a fixed component of the proposed framework; other forecasting architectures can be evaluated within the same processing, sequence-construction, and evaluation workflow.

\subsection{Envisioned Trustworthiness-Aware Evaluation}
\label{subsec:reliability-evaluation}

The sixth stage represents an envisioned extension of the current
workflow in which forecasting performance is examined in relation to
the source- and data-level characteristics associated with the
observations used to construct each trajectory sequence. The objective
is to complement conventional forecasting metrics with contextual
information describing the conditions under which the corresponding
model inputs were produced.
Let $ E_i $, denote the forecasting error associated with trajectory sample $i$, and let
\begin{equation}
    \mathbf{q}_i
    =
    \left[
    q^{\mathrm{temporal}}_i,
    q^{\mathrm{spatial}}_i,
    q^{\mathrm{density}}_i,
    q^{\mathrm{vessel}}_i
    \right]
\end{equation}
represent data-level contextual descriptors associated with that
sample. The envisioned analysis considers the general relationship
\begin{equation}
    E_i = g(\mathbf{q}_i),
\end{equation}
where $g$ is not assumed to represent a causal relationship, but
rather an empirical association between observation characteristics
and predictive performance. As the framework is extended, the
descriptor vector can additionally incorporate source-level
characteristics such as provenance, coverage, acquisition mechanisms,
and known limitations.

Three complementary forms of analysis are envisioned. First,
trajectories can be stratified according to temporal reporting
characteristics, allowing prediction errors to be compared across
different reporting conditions. Second, forecasting errors can be
aggregated spatially using the same HEALPix representation employed
for data characterization. For cell $h$, the mean prediction error
can be expressed as
\begin{equation}
    \bar{E}_h =
    \frac{1}{|\mathcal{S}_h|}
    \sum_{i \in \mathcal{S}_h}
    E_i,
\end{equation}
where $\mathcal{S}_h$ denotes the set of forecast samples associated
with cell $h$. Comparing $\bar{E}_h$ with reporting descriptors such
as $G_h$ would enable geographical associations between observation
characteristics and forecasting performance to be examined. Third,
forecasting results can be analysed across broad vessel categories
to investigate whether different maritime operating profiles are
associated with differences in observation characteristics and model
behaviour.

Forecast accuracy can be evaluated using trajectory-based distance
metrics. For an observed future trajectory $\mathbf{Y}$ and predicted
trajectory $\hat{\mathbf{Y}}$, the Average Displacement Error (ADE) is
defined as

\begin{equation}
    \mathrm{ADE}
    =
    \frac{1}{H}
    \sum_{k=1}^{H}
    d
    \left(
    \mathbf{p}_{t+k},
    \hat{\mathbf{p}}_{t+k}
    \right),
\end{equation}

while the Final Displacement Error (FDE) is

\begin{equation}
    \mathrm{FDE}
    =
    d
    \left(
    \mathbf{p}_{t+H},
    \hat{\mathbf{p}}_{t+H}
    \right).
\end{equation}

The envisioned extension would complement conventional forecasting  evaluation with contextual analysis of the observations underlying each prediction. 
This would provide two complementary
views of model performance: \emph{how accurately the model predicts}
and \emph{under what observation conditions those predictions are
produced}. Together with the source and data-characterization stage,
this establishes a concrete direction for extending the current
\emph{damast}-based workflow toward trustworthiness-aware geospatial
AI.

\section{Experimental Evaluation}
\label{sec:experiments}

This section demonstrates the implemented components of the
\emph{damast} based maritime geospatial AI workflow \cite{2026:Roehr:damast-sigspatial2026} using AIS observations from multiple public data providers.
The objective is to illustrate the reproducible path from acquisition of public AIS data through metadata-aware processing and trajectory preparation to machine-learning experimentation. In accordance with
Figure~\ref{fig:source-aware-framework}, the experiments focus on
Stages~1, 2, 4, and 5 of the workflow; the explicit source- and
data-characterization and trustworthiness-aware evaluation introduced in Stages~3 and 6 remain envisioned extensions.

The experiment is designed as a reproducible case study rather than as a benchmark of forecasting architectures. Processing pipelines, their parameterization, and machine-learning experiment
configurations are represented as reusable artifacts, allowing the
same workflow to be inspected, reproduced, and extended with
alternative predictive models. The implementation and replication package for the case study are publicly available in the accompanying GitHub repository~\cite{2026:Roehr:damast-sigspatial2026}.


\subsection{Public AIS Dataset and Experimental Setup}
\label{subsec:dataset}
The experimental dataset used for this paper consists of publicly available AIS
observations obtained from BarentsWatch ~\footnote{https://barentswatch.no}, Danish Maritime Authority~\footnote{https://www.dma.dk} and Marine Cadastre for 1 June 2026 ~\footnote{https://hub.marinecadastre.gov/pages/vesseltraffic}. The selected providers represent public AIS infrastructures covering different geographical regions and data-access settings, allowing the workflow to demonstrate harmonization and processing across heterogeneous
public AIS sources. We deliberately restrict the case study to a single day in order to provide a compact and reproducible demonstration of the processing and experimentation workflow, rather than to claim comprehensive coverage of  maritime traffic.

Data retrieved from the public data sources are converted to Parquet. During conversion,
\emph{damast} associates the dataset with metadata describing the
columns and their units, together with file-specific value ranges and descriptive statistics. The resulting metadata remain accessible alongside the observations and can be inspected programmatically or through the \emph{damast} command-line interface. An initial data pipeline merges the used data sources, while harmonizing the column format for further processing. 
The merged input contains 30,208,330 AIS observations associated with 34,376 unique MMSI identifiers. Applying the case-study ship-type filter (AIS type codes 30, 60, 70, and 80) retains 2,663,705 observations for subsequent processing. Spatial indexing is performed using HEALPix with $n_{\mathrm{side}}=128$. These concrete parameters are preserved in the serialized processing pipeline provided with the reproduction package \cite{2026:Roehr:damast-sigspatial2026}.



\begin{lstlisting}[language=Python, caption=Using damast's Python API]
pipeline = (
  DataProcessingPipeline(name=args.pipeline_name, base_dir=base_dir)
    .add("parse_timestamp", ParseTimestamp(),
         name_mappings={"from": "msgtime", 
           "to": "timestamp"})
    .add("filter_ship_types", 
         FilterWithin(within_values=args.ship_types),
         name_mappings={"x": "shipType"})
    .add("delta_time", AddDeltaTime(),
         name_mappings={"group": "mmsi", 
            "time_column": "timestamp"})
          ...
    .add("healpix", Healpix(nside=args.healpix_nside))
\end{lstlisting}

Each transformation explicitly specifies its required inputs,
including representation types, units, and admissible value ranges,
together with the outputs generated by the transformation. These
specifications make the assumptions associated with individual
processing stages explicit and allow compatibility between successive
pipeline elements to be validated.

\begin{lstlisting}[language=Python, caption=Transform lat/lon to healpix]
class Healpix(PipelineElement):
    """
    Add a HEALPix pixel id (nested scheme) per lat/lon position.
    """
    nside: int
    def __init__(self, *, nside: int = 128):
        self.nside = nside

    @damast.core.describe("Add healpix id")
    @damast.core.input({
        "lat": {"representation_type": float, "unit": "deg"},
        "lon": {"representation_type": float, "unit": "deg"},
    })
    @damast.core.output({"healpix_id": {"representation_type": int, "description": "Healpix location"}})
    def transform(self, df: AnnotatedDataFrame) -> AnnotatedDataFrame:
        latitude = self.get_name("lat")
        longitude = self.get_name("lon")
\end{lstlisting}

The resulting processing pipeline can be exported as a reusable
artifact that preserves its concrete configuration. The exported
representation can subsequently be applied to compatible datasets,
provided that \emph{damast} and the required transformation plugins
are available. This enables both the processing logic and its
parameterization to be shared rather than requiring the workflow to
be reconstructed from individual scripts. For workflows involving
multiple data sources with different preparation paths, the
processing infrastructure additionally provides mechanisms for
joining the resulting representations.




\subsection{Trajectory Forecasting: Model Creation}
\label{subsec:forecasting-results}
We illustrate the model-agnostic experiment design of \emph{damast}
for vessel trajectory forecasting. A \texttt{ForecastTask} defines
the input features, sequence and forecast lengths, preprocessing
pipeline, and training configuration while leaving the predictive
model configurable by the experimenter. Listing~3 illustrates this
interface using alternative model instances; in the case study
described in this paper, the predictive component is instantiated
with the Transformer-based forecasting model introduced in
Section~\ref{subsec:predictive-modelling}. The task constructs fixed-length, vessel-specific sequences by grouping observations by MMSI and interpolating the available observations onto equidistant time points within the configured temporal window; the forecasting horizon is specified independently of the historical sequence length. The forecasting data are split by MMSI, such that observations from a given MMSI are assigned to only one data split, preventing vessel-level leakage between the training, validation, and test sets.
\begin{lstlisting}[language=Python, caption=Defining a learning task using damast's Python API]
forecast_task = TemporalForecastTask(
  label="forecast-ais-short-sequence",
  pipeline=pipeline, features=features,
  models=[
    ModelInstanceDescription(BaselineTransformer, {}),
    ModelInstanceDescription(BaselineLSTM, {})],
  group_column="mmsi", timestamp_column="timestamp",
  sequence_length=5, window="30m",
  forecast_horizon="5m", forecast_length=1,
  max_gap="5m",
  training_parameters=TrainingParameters(
      epochs=100, validation_steps=20)
)
\end{lstlisting}

During model development, alternative predictive models can be compared while retaining the same preprocessing and sequence-construction pipeline. An \texttt{Experiment} comprises (a) data preparation using the specified processing pipeline and (b) training and evaluation of the configured forecasting models. 
Similarly to a pipeline, an experiment will generate an artifact for later introspection, preserving references to the processed data source(s), the processing pipeline, model configuration, and experiment performance results.

\begin{lstlisting}[language=Python, caption=Defining an Experiment using damast's Python API]
experiment = Experiment(learning_task=forecast_task,
                        input_data=dataset_filenames,
                        output_directory=tmp_path)
report = experiment.run()
\end{lstlisting}

Users can decide to track pipeline execution and experiment training by integrating with mlflow~\footnote{\url{https://mlflow.org}}.
The present case study focuses on the reproducible construction and configuration of the forecasting workflow rather than on benchmarking forecasting accuracy. Quantitative comparison using trajectory-error metrics such as ADE and FDE, including comparison with kinematic baselines such as constant-velocity prediction, is left for future evaluation.


\section{Discussion}
\label{sec:discussion}

The case study highlights two complementary aspects of trustworthy geospatial AI: reproducible analytical workflows and explicit consideration of the characteristics of the underlying data. The implemented \emph{damast}-based workflow addresses the first by preserving data semantics, processing transformations, and machine-learning experiment configurations, providing a traceable path from public AIS observations to predictive experiments.
The envisioned source and data-characterization and
trustworthiness aware evaluation stages address the second aspect
by relating predictive performance to observation characteristics such as provenance, reporting regularity, spatial coverage, observation density, and missingness. Together, these components provide a path from reproducible geospatial AI toward more explicit trustworthiness-aware evaluation.

\subsection{Implications for Public Geospatial AI}
\label{subsec:implications}

Public availability does not imply uniform suitability of data for
geospatial AI. Observations may vary across space, time, and observed entities, influencing the representations supplied to predictive models. The reproducible workflow demonstrated in this study preserves how such observations are transformed into model-ready
trajectories, while the envisioned source and data-characterization
stages make properties such as reporting regularity, observation
density, spatial coverage, and vessel-related variation explicit.
Relating these characteristics to downstream model performance
would enable predictions to be interpreted in the context of the
observations from which they are derived, providing a more
transparent basis for geospatial AI using heterogeneous public data.

\subsection{Reproducibility and Trustworthiness}
\label{subsec:trustworthiness}

Reproducibility contributes to trustworthiness because geospatial AI
predictions depend not only on the trained model, but also on data
interpretation, preprocessing, feature construction, sequence generation,
and experimental configuration. The damast-based implementation
maintains metadata, reusable processing pipelines, and machine-learning
experiment configurations, providing traceability from public AIS
observations to predictive outputs. The envisioned source and data characterization and trustworthiness-aware evaluation stages would extend this traceability by allowing model performance to be examined in relation to the
observation conditions under which the input data were produced. In this sense, reproducibility provides a foundation for, rather than
a substitute for, broader trustworthiness considerations.



\subsection{Industrial and Operational Relevance}
\label{subsec:industrial-relevance}

The potential industrial impact of the \emph{damast}-based workflow
and its envisioned trustworthiness-aware extension lies in supporting
more transparent and maintainable maritime AI pipelines for applications
such as vessel-traffic monitoring, route planning, fleet analytics,
and maritime situational awareness. In operational
settings, predictive accuracy alone may be insufficient: practitioners
also need information about the observation conditions underlying a
prediction. The explicit representation of the data-processing pipeline provides
a basis for incorporating source- and data-level characteristics, such
as provenance, reporting regularity, and spatial observation patterns,
alongside predictive outputs. Extending the workflow in this direction
could help practitioners identify observation conditions under which
model outputs may require additional scrutiny.
Moreover, its model-agnostic design and reproducible damast-based
processing allow predictive components to be
compared or replaced without rebuilding the complete analytical
workflow, supporting the transition from experimental models toward
reusable and maintainable maritime geospatial AI systems.
\subsection{Limitations}
\label{subsec:limitations}

The present study identifies relevant observable characteristics of public AIS data for future systematic characterization, including reporting regularity, spatial patterns, observation density, and vessel categories. These characteristics do not capture all factors
affecting data availability or quality, and source-level properties such as provenance, acquisition mechanisms, and coverage limitations are not yet represented systematically. 
Systematically quantifying how reporting gaps, observation density, spatial coverage, and other source and data-level characteristics relate to forecasting performance remains part of the envisioned extension and requires evaluation over longer temporal periods and broader observation conditions. Finally, the experimental study is restricted to a single day of AIS observations from the selected public data providers generalization across longer temporal periods, additional geographical regions, and different acquisition settings remains to be investigated.

\label{sec:conclusion}
\section{Conclusion}
In this paper, we presented an approach toward bridging the gap
between research workflows and industrial applications of geospatial
AI. We developed \emph{damast} to support reproducible data processing
and machine-learning experimentation through metadata-aware,
reusable processing pipelines.
Building on this implemented foundation, we envision extending the
workflow toward explicit source and data-characterization and
trustworthiness-aware evaluation.
By retaining reporting, spatial, and vessel-related characteristics during evaluation, such an extension would enable forecasting performance to be interpreted in relation to the observation conditions from which predictions are produced.
Future work will focus on implementing the envisioned source- and
data-characterization and trustworthiness-aware evaluation stages,
and on extending the reproducible \emph{damast} workflow to additional
GeoAI tasks, including AIS anomaly detection and trajectory forecasting
across multiple public geospatial data sources.

\section{Acknowledgments}
This work is funded by the European Commission through the project AI4COPSEC (Boosting EU Copernicus Security and Maritime Monitoring with AI and Machine Learning (ML), grant agreement No 101190021).


\printbibliography

\end{document}